# Multistep Electric Vehicle Charging Station Occupancy Prediction using Hybrid LSTM Neural Networks

**Tai-Yu Ma**\*, Luxembourg Institute of Socio-Economic Research (LISER), 11 Porte des Sciences, 4366 Esch-sur-Alzette, Luxembourg

Email: tai-yu.ma@liser.lu; ORCID: 0000-0001-6900-098X

**Sébastien Faye**, Luxembourg Institute of Science and Technology (LIST), 5, avenue des Hauts-Fourneaux, 4362 Esch-sur-Alzette, Luxembourg

Email: sebastien.faye@list.lu

\*Corresponding author

## Abstract

Public charging station occupancy prediction plays key importance in developing a smart charging strategy to reduce electric vehicle (EV) operator and user inconvenience. However, existing studies are mainly based on conventional econometric or time series methodologies with limited accuracy. We propose a new mixed long short-term memory neural network incorporating both historical charging state sequences and time-related features for multistep discrete charging occupancy state prediction. Unlike the existing LSTM networks, the proposed model separates different types of features and handles them differently with mixed neural network architecture. The model is compared to a number of state-of-the-art machine learning and deep learning approaches based on the EV charging data obtained from the open data portal of the city of Dundee, UK. The results show that the proposed method produces very accurate predictions (99.99% and 81.87% for 1 step (10 minutes) and 6 steps (1 hour) ahead, respectively, and outperforms the benchmark approaches significantly (+22.4% for one-step-ahead prediction and +6.2% for 6 steps ahead). A sensitivity analysis is conducted to evaluate the impact of the model parameters on prediction accuracy.

## 1. Introduction

Electric vehicles (EVs) have been promoted as a widely accepted solution to reduce global $CO_2$ emissions and climate change. To make low-emission energy alternatives widely accepted, charging and maintenance infrastructure needs to be widely available across Europe (Eickhout, 2017). While there has been an increase in recharging facilities installed in different countries, there is still a limited number of fast/rapid chargers (also referred to as charging points) due to their high investment cost. For example, at the end of 2020, there were only 51 public charging points in Manhattan, New York City (51 Level 2 and 0 Level 3 chargers).[1] The limited number of fast/rapid public chargers has become one of the major obstacles for widespread EV adoption (Engel et al., 2018). As there are more and more EVs running on the road, it becomes a struggle to find a charging point before running out of battery. While existing EV station platforms like ChargePoint ([www.chargepoint.com](www.chargepoint.com)) or ChargeHub ([https://chargehub.com](https://chargehub.com)) provide real-time charging point availability information for users, reservation in advance on public charging stations is still not available (Sawers, 2019). EV users might end up waiting in a queue when arriving at a charging station that was unoccupied a couple of minutes previously. A recent study shows that operating a fleet of EVs for transport network companies brings additional challenges as EVs need to charge several times a day and primarily rely on fast/rapid chargers (Jenn, 2019). Mitigating congestion at public fast/rapid charging stations has become an important issue for the efficiency of charging infrastructure management and for improving overall user experience and EV acceptance by the general public.

For this purpose, predicting charging occupancy patterns allows charging service platforms to better manage the limited charging resources available and reduce a customer's charging waiting time loss. For example, with predicted charging waiting times at charging stations, a real-time vehicle–charging station assignment/recommendation system could be developed to reduce the charging waiting time of EV fleets (Tian et al., 2016; Yuan et al., 2019; Ma and Xie, 2021; Ma, 2021). It can also support the development of apps for reducing vehicle idle time when terminating charging sessions (EU Science Hub, 2019) that are directly integrated into the vehicle's user interface (with cellular connectivity) or on the user's smartphone. These smart charging service management applications rely on accurate EV charging pattern predictions for a short-term time horizon (e.g. from 30 minutes to several hours). However, there are still few studies that address this issue, and they are mainly based on conventional econometric or time series methodologies with limited accuracy (Bikcora et al., 2016; Motz et al. 2021; Soldan et al. 2021).

Modelling the discrete EV charging occupancy state (i.e. a charger is occupied or not at a discrete-time index) at charging points is challenging as the target-dependent variable is a non-stationary binary time series, and there is very limited information available on public charging datasets (Amara-Ouali et al., 2020). Figure 1 illustrates an example of such discrete charging occupancy states over each discrete time interval (10 minutes) of a rapid charger in the City of Dundee, UK. Irregular within-day and day-to-day charging occupancy patterns can be observed. In such a problem, it is not difficult to incorporate previous occupancy states to predict the occupancy state at the next time step. However, it is more complex to consider long-term tendencies and to predict the charging occupancy sequence over multiple time steps. In general, public EV charging session data contains limited information such as the start and end times of charging sessions, the amount of energy charged of each session, the charger type or power, the geographical location of the charger, and the charger or customer ID.

---

[1] Manhattan, New York EV Charging Stations Info. [https://chargehub.com/en/countries/united-states/new-york/manhattan.html](https://chargehub.com/en/countries/united-states/new-york/manhattan.html)

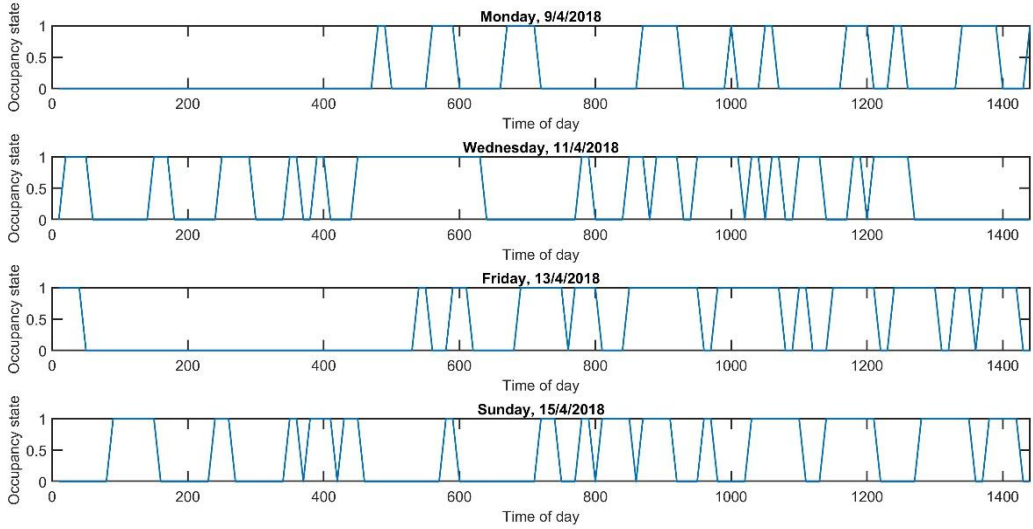

Figure 1. Example of the charging occupancy profile of a rapid charger on different days of the week (0: unoccupied, 1: occupied by an EV).

With more EV charging data made available publicly (Lee et al., 2014; Flammini et al., 2019; Amara-Ouali et al., 2020), several scholars have started modelling EV charging patterns (Iversen et al., 2014; Bikcora et al., 2016). Amara-Ouali et al. (2020) provided a complete list of 60 publicly available EV charging datasets relevant for EV charging load modelling. The available data fields in these datasets are limited, as previously mentioned. Given the variables for which data is available, the occupancy and charging load can be calculated. Two categories of problems are generally studied using these public charging datasets.

The first category concerns modelling the charging load of individual EVs and charging stations. The objective is to predict the charging load profiles to evaluate the impact on the power grid or to develop algorithms for smart charging management (Lee et al., 2019). Existing studies mainly apply statistical models to estimate the probability distribution of charging loads. Majidpour et al. (2016) applied machine learning approaches to predict aggregate charging loads at EV charging stations. Lee et al. (2019) applied Gaussian mixture models to estimate the distributions of charging arrivals, the duration of each charging session, and the amount of energy charged. Flammini et al. (2019) estimated a mixture of multivariate normal distributions for the distribution of the number of charging sessions on the public charging network in the Netherlands.

The second category considers the problem of modelling and predicting the charging occupancy profile at chargers. The outcome allows estimating energy demand together with charging power or designing algorithms to assign EVs to chargers with the least charging waiting time loss (Ma et al., 2019; Pantelidis et al., 2020). For example, Gruoss et al. (2020) proposed a Markov chain model to model the occupancy state of charging stations based on data collected for a fleet of car-sharing vehicles on about 40 public charging stations. However, the prediction accuracy of the proposed model is not reported. Bikcora et al. (2016) applied an autoregressive logistic model for day-ahead charging station availability and charging rate prediction. Iversen et al. (2014) proposed a state-space model to predict the status of EVs (driving or idle) over multiple discretised time steps. Verma et al. (2019) analysed the load profile of household energy consumption and applied different classification methods, including classification and regression trees, random forest, and k-nearest neighbour methods, to identify the energy consumption profiles of households with or without EV charging. Motz et al. (2021) applied the logistic regression model to predict the charging station occupancy using ACN charging data (Lee et al., 2019). Soldan et al. (2021) also applied the logistic regression model to predict the charging station occupancy using time-related and historical occupancy as input features. Existing studies are mainly based on econometric or machine learning approaches for charging load and occupancy prediction.

Recently, deep learning (DL) approaches have received increasing interest and have been successfully applied in speech recognition, natural language translation, computer vision, and medical image analysis, among many other fields. In particular, the long short-term memory (LSTM) network and its variants have been successfully applied in various time series forecasts (see the recent review in Van Houdt et al., 2020). An LSTM network is a kind of recurrent neural network allowing the modelling of complex temporal dependency in time series data and overcoming the vanishing gradient problem. In the energy field, DL has been successfully applied to residential load profile forecasting (Kim and Cho, 2019; Yang et al., 2019; Sajjad et al., 2020), energy consumption prediction (Ullahet al., 2019; Wang et al., 2020), and gas consumption profile prediction (Laib et al., 2019). However, the problems considered in these studies are regression problems. There are still few studies that tackle discrete EV charging occupancy state modelling problems.

In this study, we propose a hybrid LSTM neural network that combines both LSTM and forward neural networks to merge heterogeneous features for the prediction of EV charging occupancy over a planning horizon. The results show that the proposed approach outperforms the state-of-the-art machine learning approaches and benchmark deep learning networks using the public charging data from the City of Dundee, UK, in 2018. The main contributions of this paper are summarized as follows.

- We develop a new hybrid LSTM method to predict discrete EV charging occupancy sequences for multistep prediction. We generate new day-type tendency features from limited fields of public EV charging station data to increase significantly the prediction accuracy. The proposed mixed network structure allows merging heterogeneous data types, providing a generalized and flexible framework for time series forecasting based on LSTM neural networks.
- Our experiments show that the proposed approach outperforms significantly classical time series (logistic regression), machine learning approaches (support vector machine (SVM), random forest, and Adaboost), and benchmark deep learning networks (LSTM, Bi-LSTM, and GRU). The proposed method is easy to implement and can be applied for efficient charging infrastructure management.
- We also analyze the EV charging session data of the city of Dundee to understand the charging patterns of users and charging load profiles in the study area and evaluate the influence of different hyperparameters of the hybrid LSTM method on prediction accuracy.

This paper is organized as follows. Section 2 presents the dataset and the exploratory analysis of the data. Section 3 presents the proposed hybrid LSTM model, performance metrics, and alternative benchmark methods for multistep EV charging occupancy state prediction. In Section 4, we report the performance of the proposed method and compare it with the benchmark methods. A sensitivity analysis is conducted to evaluate to what extent different model parameters affect the prediction accuracy for multistep prediction. Finally, we discuss the results and offer some concluding remarks.

## 2. Data collection and pre-processing

### 2.1. Dataset

In this paper, we consider EV charging data from the open data portal of the city of Dundee, UK,[2] which provides datasets describing various EV charging sessions. Each session contains charger identifiers, start and end times of charging sessions, amount of energy charged, the power of chargers, and the geographical locations of chargers. There are three types of chargers in the study area: slow chargers (7kW), fast chargers (22kW), and rapid chargers (≥43kW).[3]

In the present study, a three-month charging session dataset from March 5, 2018, to June 4, 2018 (91 days) is used. Table 1 reports the descriptive statistics. There is a total of 40 slow, 8 fast, and 9 rapid chargers in the data; 56.2% of charging sessions are realized at rapid chargers, while 35.2% and 8.5% are at the slow and fast chargers, respectively. Due to the low charging occupancy of slow and fast chargers (1.5 and 1.9 sessions per day-charger for slow and fast chargers, respectively), we focus on the more challenging issue of modelling the occupancy of rapid chargers (10.9 sessions per day-charger, on average), which present very irregular within-day and day-to-day occupancy patterns as shown in Figure 1.

---

[2] https://data.dundeecity.gov.uk/dataset/ev-charging-data
[3] https://www.zap-map.com/charge-points/connectors-speeds/

We first delete the outliers: charging durations at rapid chargers that are more than three standard deviations from the median (Verma et al., 2019). A total of 0.79% outliers are removed. Ultimately, 8870 rapid charger charging sessions are used for this study. The average charging duration of the rapid chargers is 28.3 minutes, with a standard deviation of 36.4 minutes (see Table 1). The geographical locations of the charging stations in the study area are shown in Figure 2.

Table 1. Descriptive statistics of EV charging sessions in the study area.

| Charger type | Number of charging points | Number of charging sessions | Number of charging sessions per day per charging point | Charging duration (minutes) | |
|---|---|---|---|---|---|
| | | | | Mean | S.D. |
| Slow (7kW) | 40 (70.2%) | 5603 | 1.5 | 725.8 | 1338.5 |
| Fast (22kW) | 8 (14.0%) | 1357 | 1.9 | 413.3 | 819.9 |
| Rapid (43kW or above) | 9 (15.8%) | 8941 | 10.9 | 28.3 | 36.4 |
| Total | 57 (100%) | 15901 | 3.1 | 306.9 | 892.0 |

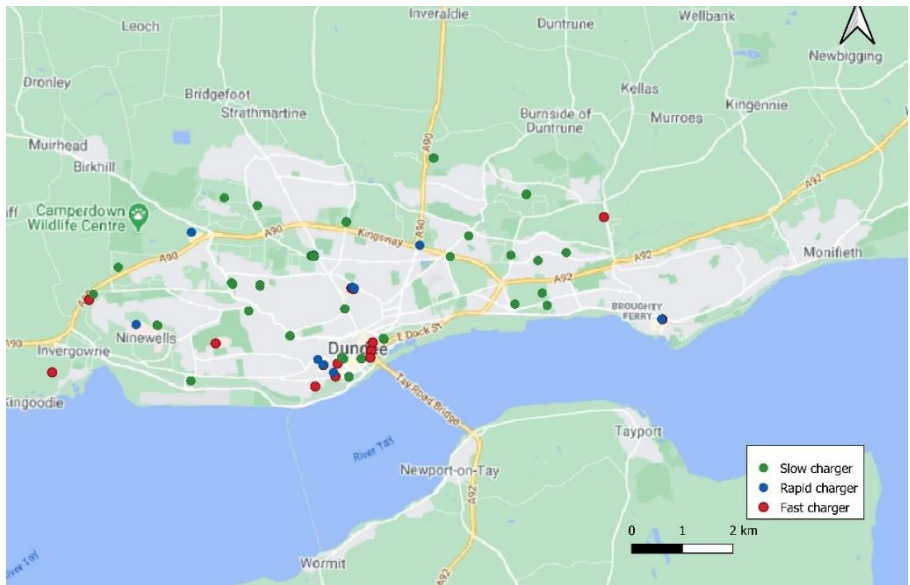

Figure 2. The spatial distribution of chargers in the city of Dundee *(source: https://data.dundeecity.gov.uk/dataset/ev-charging-data)*.

*2.2. Characteristics of charging occupancy at rapid chargers*

We further analyse the charging behaviour of users and the charging occupancy patterns at rapid chargers. The upper part of Figure 3 shows the EV plug-in time distributions on weekdays and weekends. On weekdays, most charging sessions occur from 7:00 to 21:00, with a peak between 12:00 and 14:00. For weekends, the plug-in time profile is smoother compared to that of weekdays. The number of overnight (from 0:00 to 6:00) charges are doubled on the weekend. Regarding the charging duration on weekdays and weekends (the middle part of Figure 3), almost all charging sessions are less than 60 minutes, and mostly between 10 and 40 minutes. The lower part of Figure 3 shows the distribution of the charged energy on weekdays and weekends. The charging occupancy profiles are irregular and present a significant difference between weekdays and weekends, as shown in Figure 4.

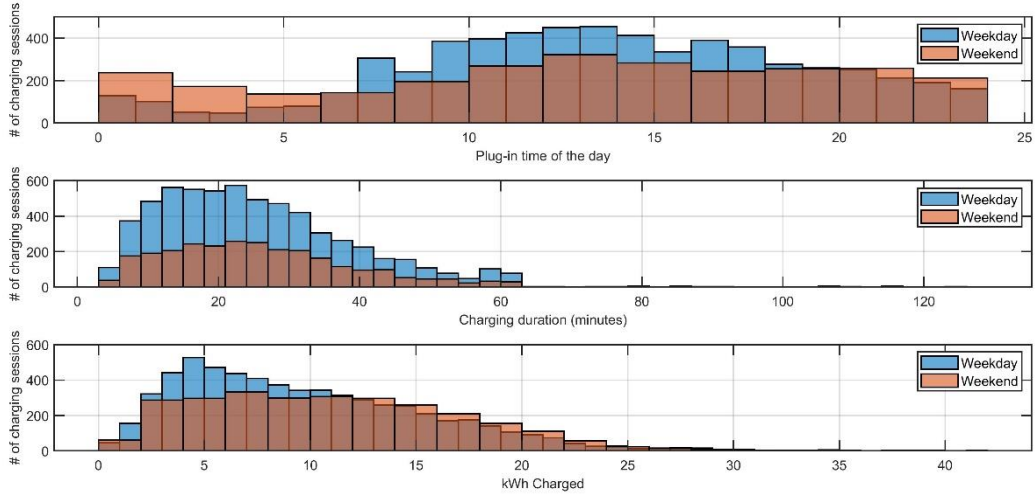

Figure 3. Distribution of EV plug-in times, charging duration, and energy charged using rapid chargers on weekdays and weekends (March 5, 2018, to June 4, 2018).

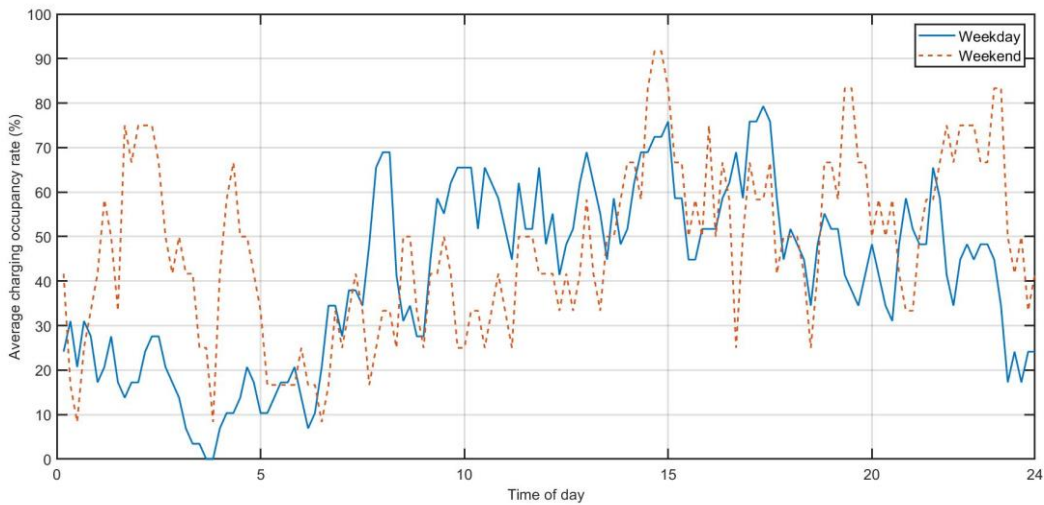

Figure 4. Example of the average charging occupancy rate on weekdays and weekends.

*2.3. Factors influencing charging occupancy profiles*

The occupancy profile of chargers depends on various factors, including the time of the day, day of the week, weekday/weekend, charging power, remaining battery level of EVs, battery capacity, energy price, the geographical location of charging stations, and idle time during which a vehicle is fully charged but still occupies a charger (EU Science Hub, 2019). Given the limited data fields available in our datasets, the features generated for modelling the charging occupancy state include the time of day, day of the week, weekend, and the past charging occupancy states and average charging occupancy rates on the type of the day (weekday or weekend), as shown in Table 2. The objective is to predict the charging occupancy state profile for each charger for multiple steps ahead (from 10 minutes to several hours). For this purpose, we discretised one day (24 hours) into 144 discrete time slots with 10-minute intervals. The entire dataset is divided into a training dataset (first 70% of the dataset) and a test dataset (remaining 30% of the dataset). The auto- and partial correlations of charging occupancy states show that the occupancy state at time *t* is correlated with its past states at times *t–1* and *t–2* (Figure 5). Different from existing studies that consider time-related and historical occupancy state features only, we create a long-term charging occupancy tendency as an additional feature to help predict the charging occupancy state for multiple steps ahead. Figure 4 illustrates an example showing that the charging occupancy profiles are volatile and distinct on weekdays vs weekends.

Table 2. Features used to model the charging occupancy of rapid chargers.

| Feature (Variable) | Meaning |
| --- | --- |
| Time of day ($t$) | Time index belonging to $t \in \{1,2,...,144\}$ for 24 hours |
| Day of week ($d$) | Sunday=0, Monday=1, …, Saturday=6 |
| Weekday/weekend ($w$) | 1 if charging occurs on the weekend and 0 otherwise |
| Average charging occupancy rate profile for weekday/weekend ($p$) | A vector of 144 continuous variables representing the tendency of the charging occupancy rate of a charger on weekdays or the weekend. Two constant vectors are generated, one for weekdays and another for the weekend, calculated from the 70% training dataset. |
| Past charging occupancy states ($y$) | A sequence of historical k-step backward charging occupancy states from t (i.e. $t-1, t-2, ..., t-k$). |

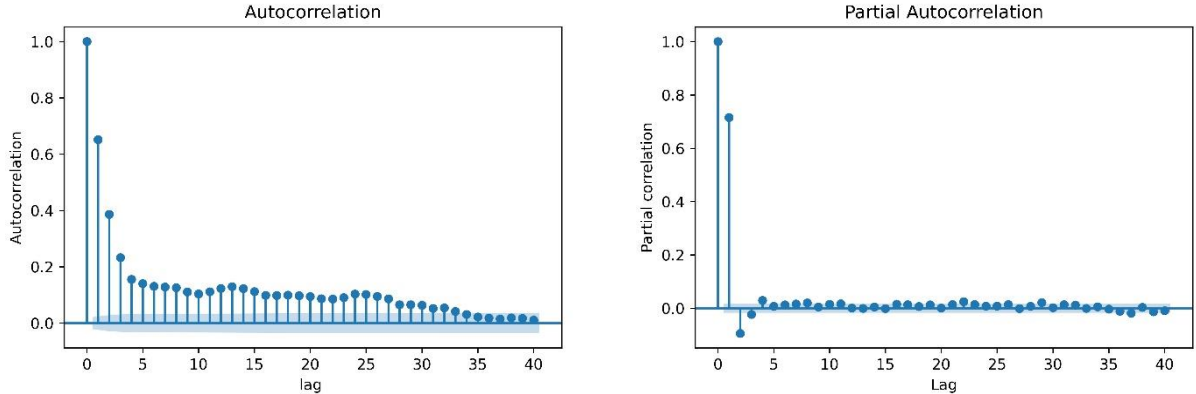

Figure 5. Example of the auto- and partial correlations of the charging occupancy states at chargers.

## 3. Occupancy state prediction models

### 3.1. Proposed hybrid LSTM model

Consider an EV charger occupancy state time series $y_t$ for a charger, where $t$ is a discrete-time index $t \in T = \{1,2,....,\frac{H}{\Delta}\}$, with $\Delta$ being the discrete time interval and H being 24 hours. Let $y_t$ be 1 if there is an EV charging event that occurs during $[(t-1)\Delta, t\Delta)$, and 0 otherwise. Given a sequence of past charging occupancy states $(y_{t-1}, y_{t-2}, ...)$ before $t$, we aim to predict the charging occupancy state sequence for a short-term time window ahead, i.e. $t+1, ..., t+k-1$.

We propose a hybrid LSTM model taking into account both local temporal dependency and long-term tendency to predict the charging occupancy state sequence. Figure 6 shows the network structure of the model. The short-term charging state dependency is modelled by an LSTM block, while time-related information and the long-term tendency of charging state profiles are handled by a multilayer feedforward neural network. Let a sample of the input time series data for one-time step $t$ be $X_t = \{X_{1t}, X_{2t}\}$, where $X_{1t} = \{y_{t-1}, y_{t-2}, \ldots, y_{t-m}\}$ and $X_{2t} = \{t, d, w, \boldsymbol{p}\}$ (Table 2). The LSTM block takes $X_{1t}$ as input and generates a vector of hidden states and then combines this with the output of the feedforward neural network layers using $X_{2t}$ as input (Figure 6). The latter uses three fully connected layers to learn complex features from $X_{2t}$. The output of the LSTM block and the fully connected layers are concatenated and followed by a fully connected layer and an output layer with $k$ neurons, each one corresponding to a charging occupancy state prediction for $t, t+1, \ldots, t+k-1$. Two dropout layers are used for regularization to avoid overfitting. Note that we also test incorporating the spatial correlation of charging occupancies at nearby charging stations (i.e. charging occupancies of nearby chargers) as input features, but the results show that this strategy did not improve the prediction accuracy. The detailed model structure is shown on the right side of Figure 6.

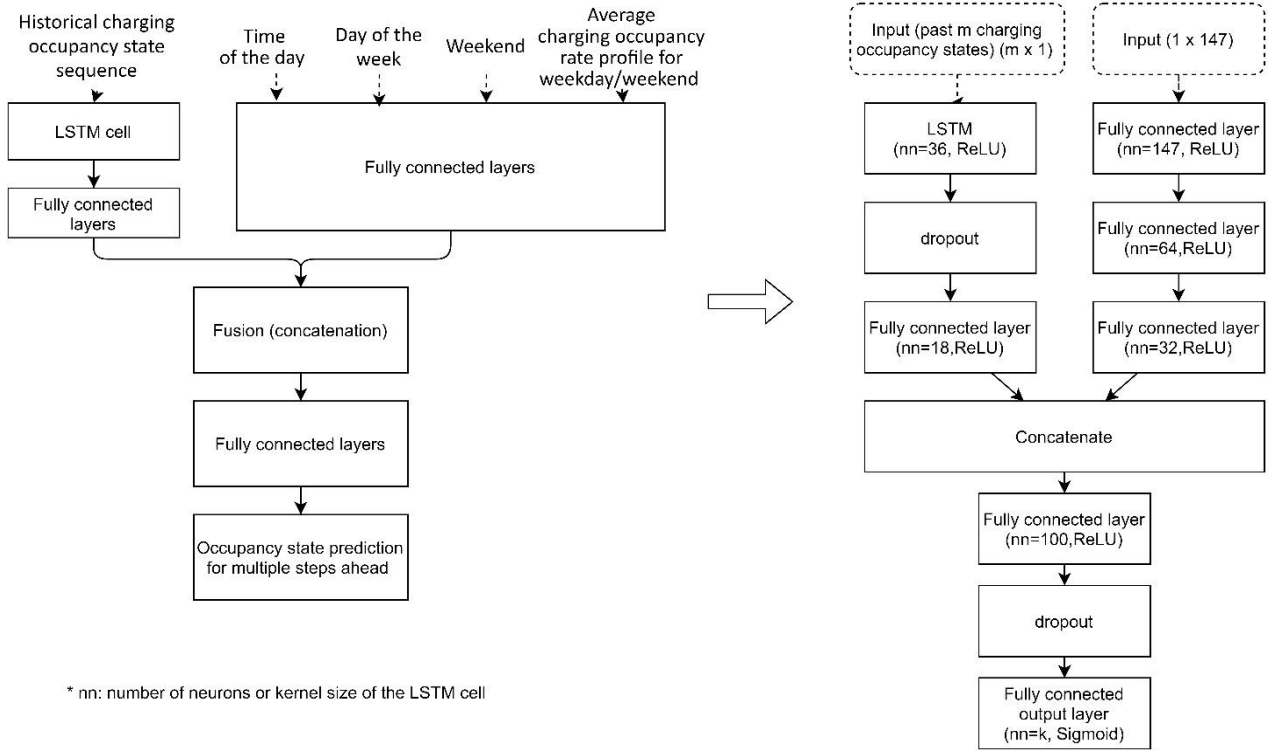

Figure 6. Proposed hybrid LSTM network architecture.

For each time step, twelve past charging states are used as input for the LSTM block, representing a 2-hour charging occupancy sequence. The choice of 12 past occupancy states is based on the experiments with different lengths of past states. Figure 7 shows the structure of the LSTM block, which contains different cells (rectangles in Figure 7), an input gate, a forget gate, and an output gate. The equations for the LSTM block are shown in Eq. (1)–(6) (Hochreiter and Schmidhuber, 1997). The input gate (Eq. (1)) combines the input vector $x_t$ ($x_t = X_{1t}$) and its hidden state vector $h_{t-1}$ at $t-1$. The forget gate (Eq. (2)) filters the long-term information to be retained. The output gate (Eq. (3)) uses a sigmoid activation function to determine what information is to be output for the cell state. Equation (4) computes a temporary cell state based on the current input and hidden state at $t-1$. The cell state at $t$ (Eq. (5)) is then updated by its previous state $c_{t-1}$ and its temporal state $\tilde{c}_t$. The output gate (Eq. (6)) controls the final output of the cell at time $t$ based on $o_t$ and $c_t$.

$$i_t = \sigma(W_{xi}x_t + W_{hi}h_{t-1} + b_i), \tag{1}$$

$$f_t = \sigma(W_{xf}x_t + W_{hf}h_{t-1} + b_f), \tag{2}$$

$$o_t = \sigma(W_{xo}x_t + W_{ho}h_{t-1} + b_o), \tag{3}$$

$$\tilde{c}_t = tanh(W_{xc}x_t + W_{hc}h_{t-1} + b_c), \tag{4}$$

$$c_t = f_t \otimes c_{t-1} + i_t \otimes \tilde{c}_t, \tag{5}$$

$$h_t = o_t \otimes \tanh(c_t), \tag{6}$$

where $W_{x*}$ and $W_{h*}$ are the set of weights connecting to the input vector $x_t$ and the previous hidden state $h_{t-1}$, respectively. $b_i, b_f, b_o, b_c$ are the corresponding bias terms. $\otimes$ is the element-wise multiplication. $i_t$ is the input gate that decides what information should be retained. $f_t$ is the forget gate that decides what information is to be removed. $o_t$ is the output gate controlling the information to be moved forward to the output at this time step. $c_t$ is the updated cell-state vector at t. $h_t$ is the output of the LSTM block at time step $t$. $\sigma$ and $tanh$ are the sigmoid and hyperbolic tangent activation functions, respectively. The output of the LSTM block is followed by a dropout layer for regularization and then connected by a fully connected layer to further extract the temporal feature.

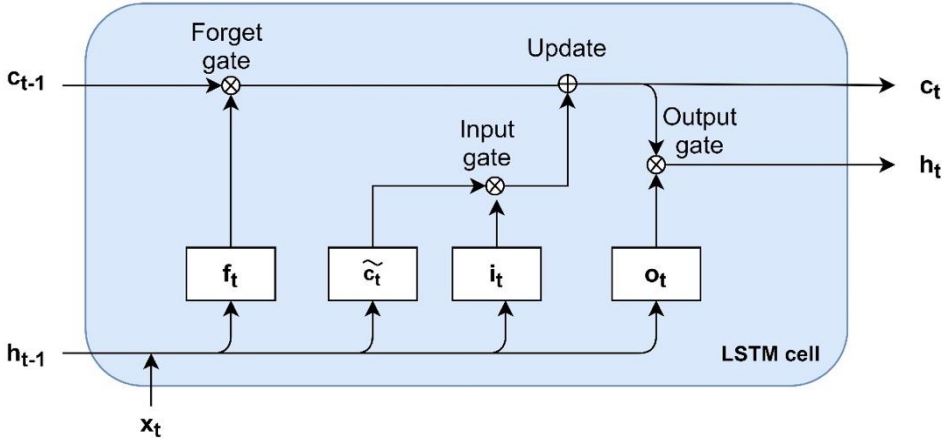

Figure 7. The LSTM block architecture.

The weights of the hybrid LSTM model are learned by backpropagation through time to minimize a designed loss function. The activation function used in the fully connected layers is defined as Eq. (7).

$$h_l = \text{ReLU}(W_l h_{l-1} + b_l), \tag{7}$$

where ReLU is defined as $\text{ReLU}(q) = max(q, 0)$. $W_l$, $h_{l-1}$, and $b_l$ are the weight vector, hidden layer vector, and bias term of layer $l$, respectively. The output of the hybrid LSTM model is a vector of binary values representing the multistep forecasting of the charging occupancy states as in Eq. (8).

$$\hat{y}_t = r(W_{ho}h_o + b_o), \tag{8}$$

where $W_{ho}$ is the hidden-to-output weights. $h_o$ is the input vector for the output layer o, and $b_o$ is the bias term. $r$ is a sigmoid function defined as $r(Z) = \frac{1}{1+e^{-Z}}$.

For hyperparameter settings, we adopt a manual coarse turning process by sequentially testing a set of hyperparameters based on the proposed hybrid LSTM network architecture (Zhang et al., 2017; Schwemmle, 2021; Schwemmle and Ma, 2021). This approach first sets up the reference values for the hyperparameters and tune a set of key hyperparameters (i.e. the size of hidden layers, kernel size of the LSTM block, number of hidden layers, dropout rate, mini-batch size, see Table 8) one by one in a sequential way. Schwemmle and Ma (2021) showed that such a hyperparameter turning strategy could obtain very accurate results with considerable training-time savings. Applying the state-of-the-art hyperparameter search algorithms (Bergstra et al., 2011) to find the best-performing hyperparameters could further achieve around a 2% to 5% performance gain, with the cost of high computational time (Schwemmle, 2021). Table 3 reports the retained final hyperparameters for the hybrid LSTM model. The total number of parameters to be trained for the final model is 45,152. The impact of the different hyperparameter settings is analysed in Section 4.3. The implementation of the hybrid LSTM model is based on Python's Keras Application Programming Interface of the TensorFlow package.

Table 3. Hyperparameter settings for the hybrid LSTM model.

| Hyperparameter | Value | Hyperparameter | Value |
| --- | --- | --- | --- |
| Size of hidden layers/LSTM block | See Figure 6 | Regularization | Dropout (0.2) |
| Number of hidden layers | See Figure 6 | Mini-batch size | 30 |
| Activation function | See Figure 6 | Number of training epochs | 15 |
| Learning rate | 0.001 | Optimizer | Adam |

*3.2. Benchmark methods and performance metrics*

In addition to the proposed hybrid LSTM model, we consider four machine-learning models as benchmarks to compare the performance of our approach. These models include logistic regression (Linoff and Berry, 2011), SVM (Smola and Schölkopf, 2004), random forest (Ho, 1995), and Adaboost (Freund and Schapire, 1997). Logistic regression uses a logistic function to model the binary outcomes (classes) based on a set of features. SVM builds a kernel-based hyperplane within a high-dimensional feature space for classification tasks. Random forest builds a number of decision-tree classifiers by sampling and averages the predicted classes from each tree to improve the prediction accuracy and avoid overfitting. Adaboost is an ensemble of learning methods that combine a set of classifiers as a weighted sum to increase prediction accuracy. The weights are adaptively adjusted to increase the accuracy of difficult misclassified cases. These supervised learning methods (classifiers) model the charging occupancy state at one time step as a binary classification problem based on a feature vector as in Eq. (9).

$$\hat{y}_t = f(X_t), \tag{9}$$

where $f$ represents an applied classifier. $X_t$ is the feature vector for time step $t$. $\hat{y}_t$ is the predicted outcome for time step $t$. The considered feature space for the benchmark classifiers is the same as in the previous section. As having irrelevant features may reduce the prediction accuracy of classifiers, we test three feature sets as follows.

− Model 1: Consider 6 features as $X_t = \{t, d, w, y_{t-1}, y_{t-2}, y_{t-3}\}$.
− Model 2: Extend Model 1 by using 12 past charging occupancy states, i.e. $X_t = \{t, d, w, y_{t-1}, y_{t-2}, \dots, y_{t-12}\}$.
− Model 3: Extend Model 2 with the additional features of average charging occupancy rates on the same type of day, i.e. $X_t = \{t, d, w, \boldsymbol{p}, y_{t-1}, y_{t-1}, \dots, y_{t-12}\}$.

Our goal is to evaluate whether there are significant performance differences when incorporating more elaborated features and determine which feature setting to use. Table 4 shows that there are no significant gains when introducing more complex features for these classifiers. As a result, the feature set of Model 1 is used for the benchmark methods to predict multistep charging occupancy states in the next section. The walk-forward approach is used for multistep time series prediction (Brownlee, 2016). This approach uses the prediction ($\hat{y}_t$) at time step $t$ as input for the prediction at the next time step $t + 1$ and moves forward for the prediction time window. The implementation of the benchmark classifiers is based on Python's Scikit-learn package.

Table 4. Prediction accuracy of the benchmark methods based on the different feature settings.

| Classifier | Model 1 (6 features) | Model 2 (15 features) | Model 3 (159 features) |
|---|---|---|---|
| Logistics | 0.8837 | 0.8833 | 0.8835 |
| SVM | 0.7511 | 0.7525 | 0.7509 |
| Random forest | 0.8370 | 0.8366 | 0.8299 |
| Adaboost | 0.8816 | 0.8818 | 0.8818 |

Remark: The results are based on the test dataset.

A loss function is defined to measure the difference between predicted and observed outcomes. For our problem, the mean absolute error (MSE) of Eq. (10) is used as an accuracy metric to measure the performance of the predicted charging occupancy sequence on a predefined short-term time window with $k$ time steps ahead.

$$MAE = \frac{1}{k}\sum_{s=t}^{t+k-1} |\hat{y}_s - y_s|, \tag{10}$$

where $y_s$ is the observed value for time step $s$ and $\hat{y}_s$ is the predicted value for time step $s$. $k$ is the length of a prediction time window. For the considered problem, the performance of a model is measured as the average accuracy moving through each time step of the test dataset. Note that as certain chargers are unoccupied for most of the time (unbalanced data), the F1 score is calculated as a complementary metric. The F1 score is a weighted measure of precision (the number of correct positive predictions divided by the total number of positive outcomes) and recall (the number of correct positive predictions divided by the sample size). Given a vector of predicted outcomes for a time window, let $TP$ denote the number of true positives (correct prediction for the outcome of 1) and $FP$ the number of false positives. The F1 score is defined as in Eq. (11).

$$\text{F1 score} = \frac{2}{precision^{-1} + recall^{-1}} = \frac{TP}{TP + \frac{FN + FP}{2}}, \tag{11}$$

where $TP$, $FN$, and $FP$ are the number of true positives, false negatives, and false positives, respectively. Note that in the case of zero division, the F1 score is set as 0.

## 4. Results analysis

### 4.1. Model performance metrics

The multistep prediction results for the hybrid LSTM model and the benchmark methods are shown in Table 5. The length of the prediction time window ranges from 1 (10 minutes) to 36 (6 hours) time steps ahead. The reported results are the average of 10 runs on the test dataset for all rapid charging stations. Figure 8 shows that the proposed hybrid LSTM model outperforms the benchmark machine learning methods significantly. The 1-step prediction of the hybrid LSTM model is very accurate (0.9999) compared to the benchmark methods (accuracy ranging from 0.7511 to 0.8837). The prediction accuracy decreases as the length of the prediction time window increases. For the prediction of 3- and 6-time steps ahead, the accuracy of the hybrid LSTM model remains satisfactory (0.8926 and 0.8187), outperforming the benchmark methods (0.8042 and 0.7563, respectively). As for the F1 score, its values drop significantly starting from 12-steps-ahead forecasting. We can conclude that the proposed approach is suitable for charging occupancy state prediction for time windows less than 60 minutes ahead.

Table 5. Performance metrics of the hybrid LSTM model and the benchmark methods.

| | Accuracy | | | | | |
|---|---|---|---|---|---|---|
| k-step ahead | 1 | 3 | 6 | 12 | 24 | 36 |
| Logistics | 0.8837 | 0.8034 | 0.7518 | 0.7166 | 0.6927 | 0.6809 |
| SVM | 0.7511 | 0.7421 | 0.7366 | 0.7328 | 0.7295 | 0.7283 |
| Random forest | 0.8370 | 0.7710 | 0.7397 | 0.7240 | 0.7162 | 0.7137 |
| Adaboost | 0.8816 | 0.8042 | 0.7563 | 0.7257 | 0.7080 | 0.6999 |
| Hybrid LSTM | **0.9999** | **0.8926** | **0.8187** | **0.7776** | **0.7593** | **0.7511** |
| | F1 score | | | | | |
| k-step ahead | 1 | 3 | 6 | 12 | 24 | 36 |
| Logistics | 0.7760 | 0.7754 | 0.6687 | 0.5423 | 0.4110 | 0.3367 |
| SVM | 0.5987 | 0.6805 | 0.6031 | 0.5003 | 0.3918 | 0.3330 |
| Random forest | 0.6860 | 0.7247 | 0.6359 | 0.5452 | 0.4566 | 0.4088 |
| Adaboost | 0.7681 | 0.7736 | 0.6676 | 0.5425 | 0.4118 | 0.3366 |
| Hybrid LSTM | **0.9999** | **0.8562** | **0.7305** | **0.6108** | **0.4896** | **0.4279** |

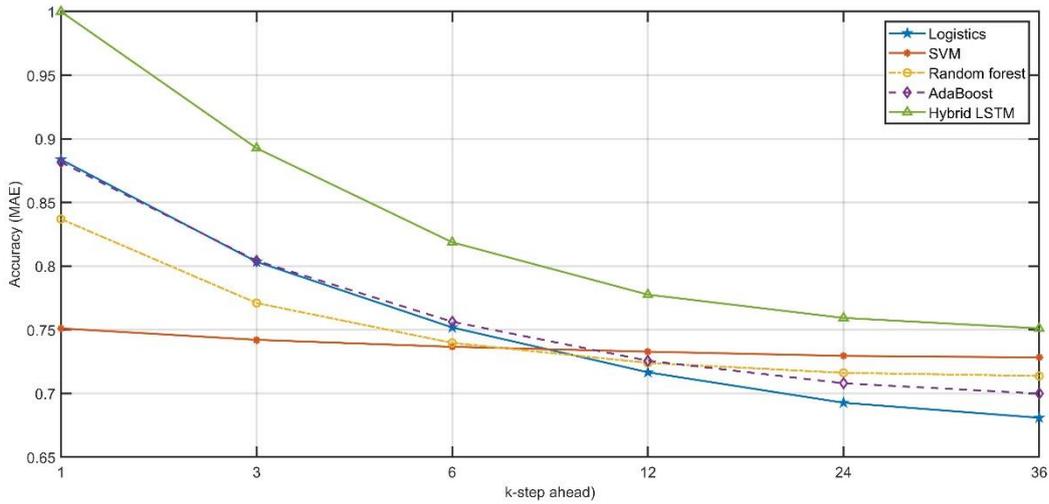

Figure 8. The prediction accuracy of the hybrid LSTM and the benchmark methods.

Table 6 shows the detailed prediction result for each rapid charger for multiple time steps ahead. The performance of the hybrid LSTM model has a good prediction accuracy of 83.7–97.9% for prediction 3-time steps ahead (30 minutes). These numbers reduce gradually when the prediction time windows become longer, as shown on the left side of Figure 9. From the right side of Figure 9, we can observe that five chargers have an average charging occupancy rate of around 40% or more, while only charger 1 has a low occupancy rate of 6.3%. As an example, Figure 10 compares the predicted and observed charging occupancy profiles for multiple time steps ahead. The results show that the proposed model predicts well the observed charging profiles over different prediction time horizons.

Table 6. Prediction accuracy of the hybrid LSTM on the test dataset for different rapid chargers.

| # of time steps ahead | Rapid charger ID | | | | | | | | |
|---|---|---|---|---|---|---|---|---|---|
| | 1 | 2 | 3 | 4 | 5 | 6 | 7 | 8 | 9 |
| 1 | 1.000 | 1.000 | 1.000 | 1.000 | 1.000 | 1.000 | 1.000 | 1.000 | 1.000 |
| 3 | 0.979 | 0.945 | 0.922 | 0.903 | 0.859 | 0.837 | 0.838 | 0.862 | 0.887 |
| 6 | 0.966 | 0.917 | 0.868 | 0.814 | 0.769 | 0.734 | 0.747 | 0.758 | 0.808 |
| 12 | 0.960 | 0.905 | 0.838 | 0.758 | 0.706 | 0.686 | 0.705 | 0.709 | 0.763 |
| 24 | 0.957 | 0.898 | 0.821 | 0.715 | 0.676 | 0.661 | 0.684 | 0.688 | 0.741 |
| 36 | 0.956 | 0.895 | 0.818 | 0.699 | 0.670 | 0.658 | 0.687 | 0.678 | 0.735 |

Remark: One-time step is 10 minutes.

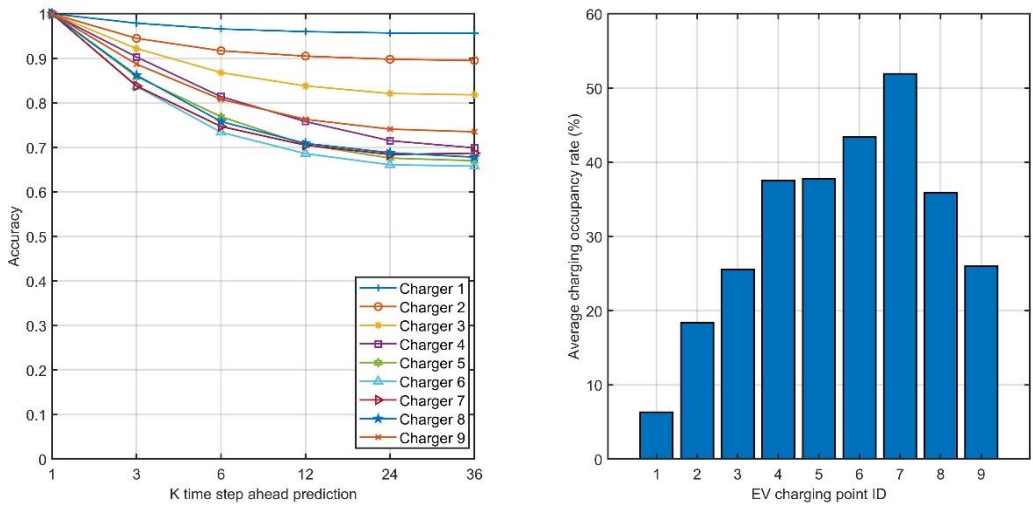

Figure 9. Prediction accuracy of the charging occupancy states for each charger over multiple time steps (left) and the average charging occupancy rates for each charger (right).

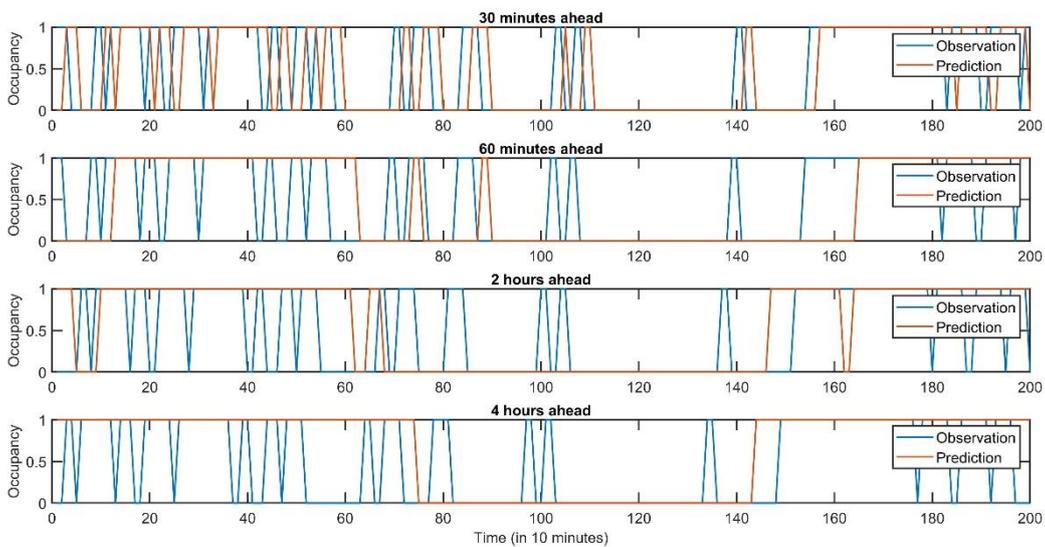

Figure 10. Example of the observed and predicted charging occupancy profiles using the hybrid LSTM approach for multiple time steps ahead.

*4.2. Performance comparison with other deep learning approaches*

We further compare the performance of the proposed hybrid LSTM model with seven benchmark DL models considering the same feature space. The benchmark DL neural network architectures are shown in Figure 11.

a. LSTM (Hochreiter and Schmidhuber, 1997): Use a classical LSTM network only to connect input sequences of feature data for multistep charging-state prediction.
b. Bi-LSTM (Kiperwasser and Goldberg, 2016): Use a bidirectional LSTM network to consider both forward and backward information of input sequences of feature data sequences for multistep prediction.
c. GRU (Cho et al., 2014): Use a gated recurrent unit (GRU) network, which is similar to LSTM but with a simpler architecture and fewer parameters to learn, for multistep charging-state prediction. For the above three network structures, the input sequences of features are connected with an LSTM/Bi-LSTM/GRU block, followed by a fully connected layer, a dropout layer, and a fully connected output layer with a sigmoid function.
d. Conv1D (Barkost, 2020): Use two 1-dimensional (1D) convolutional layers, one Max pooling layer, and one fully connected layer for sequential charging station occupancy state predictions. Input features are connected with two sequential 1D convolutional layers (kernel size=4) to filter information and then followed by a Max pooling layer. The latter is flattened and then connected by a fully connected layer and an output layer.
e. Stacked LSTM (Pascanu et al., 2013): Stack multiple LSTM layers on each other to learn deepened hidden-to-hidden state transitions for more complex pattern recognition. We connect the input features by two consecutive LSTM layers, followed by a fully connected layer, a drop-out layer, and an output layer.
f. ConvLSTM (Shi et al., 2015; Petersen et al., 2019): The convolutional LSTM is a combination of conventional networks and LSTM networks for spatiotemporally correlated data predictions by integrating convolutional filters into the LSTM structure. The input feature sequences are connected by a 2D ConvLSTM cell with a one-dimensional kernel size (1, 4) for handling one-dimensional time series data in our case. The output of the ConvLSTM cell is flattened and then connected by a fully connected layer, a dropout layer, and an output layer for multiple time-step predictions.
g. CNN-LSTM (Donahue et al., 2015): Different from the ConvLSTM, the CNN-LSTM uses multiple CNN layers to filter information and then connect their outputs by an LSTM cell for learning hidden temporal relationships. We connect the input features with multiple CNN layers and multiple Max pooling layers in-between. An LSTM cell is connected after flattening the CNN layers, then connecting to a fully connected layer, a dropout layer, and an output layer.

The feature vectors for these DL models are defined as $X_t = (V_t, V_{t-1}, ..., V_{t-m-1})$, with $V_t = \{t, d, w, \boldsymbol{p}, y_{t-1}\}$. To find well-performing feature settings, we first vary the number of historical steps *m* to find a good input sequence length and then refine the hyperparameters of the model. Finally, $X_t = (V_t, V_{t-1}, V_{t-2})$ (3 historical steps) is retained for the DL variants. The hyperparameters used for these models are shown in Figure 11, more detailed settings can be found in the computational source codes below. Table 7 compares the prediction accuracies of the hybrid LSTM model and those of the benchmark DL models. The results show that the proposed hybrid LSTM model outperforms significantly the benchmark DL models, in particular for predicting 6 or fewer time steps ahead (see Figure 12). The characteristics of the hybrid LSTM model for its outperformance can be highlighted as follows. First, time series data present different temporal regularities for which highly irregular ones are difficult to predict (e.g. customer arrival patterns and charging times might be random for certain charging stations). When applying the LSTM (or other time series/machine learning approaches) for highly irregular time series data prediction, it might achieve its maximum predictability by learning the local temporal patterns (Zhao et al., 2019). To augment the predicting accuracy, the hybrid LSTM incorporates the output of another predictor (expected charging occupancy trends on a longer horizon) using multiple fully connected layers to improve the limit of the predictability of the LSTM cells. Second, the proposed model provides a flexible framework to incorporate additional features as extended neural network branches for extracting information from different features or predictors to augment its predicting accuracy. Future extensions could consider this idea to combine predictions from other learning algorithms for better performance as is the case for ensemble learning approaches.

The data and Python codes used in this study are available at https://github.com/tym2021.

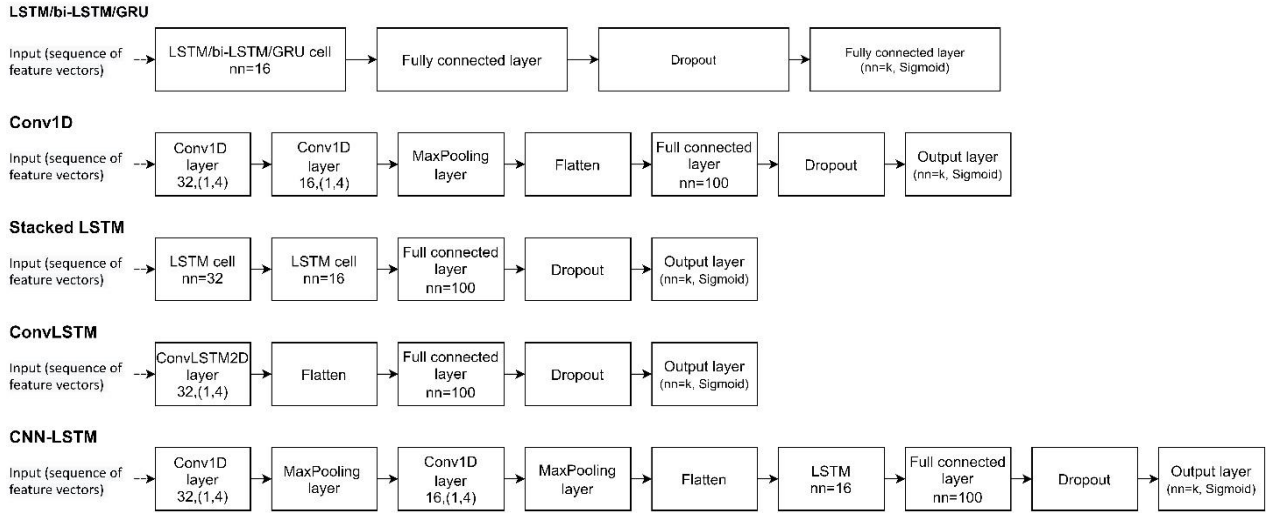

Figure 11. The benchmark DL neural network architectures (kernel size is in parentheses).

Table 7. The prediction accuracy of the benchmark DL models for multiple steps ahead.

|  | Prediction of k-time steps ahead | | | | | |
| --- | --- | --- | --- | --- | --- | --- |
| Model | 1 | 3 | 6 | 12 | 24 | 36 |
| LSTM | 0.8887 | 0.8107 | 0.7613 | 0.7528 | 0.7363 | 0.7294 |
| Bi-LSTM | 0.8888 | 0.8149 | 0.7751 | 0.7557 | 0.7351 | 0.7383 |
| GRU | 0.8835 | 0.8085 | 0.7681 | 0.7484 | 0.7414 | 0.7398 |
| Conv1D | 0.7708 | 0.7449 | 0.7388 | 0.7358 | 0.7376 | 0.7367 |
| Stacked LSTM | 0.8832 | 0.8100 | 0.7701 | 0.7504 | 0.7416 | 0.7393 |
| ConvLSTM | 0.8830 | 0.8102 | 0.7703 | 0.7488 | 0.7454 | 0.7390 |
| CNN-LSTM | 0.8834 | 0.8082 | 0.7711 | 0.7486 | 0.7450 | 0.7416 |
| Hybrid LSTM | **0.9999** | **0.8926** | **0.8187** | **0.7776** | **0.7593** | **0.7511** |

Remark: The results are based on the average of 10 runs for the test dataset of all rapid chargers.

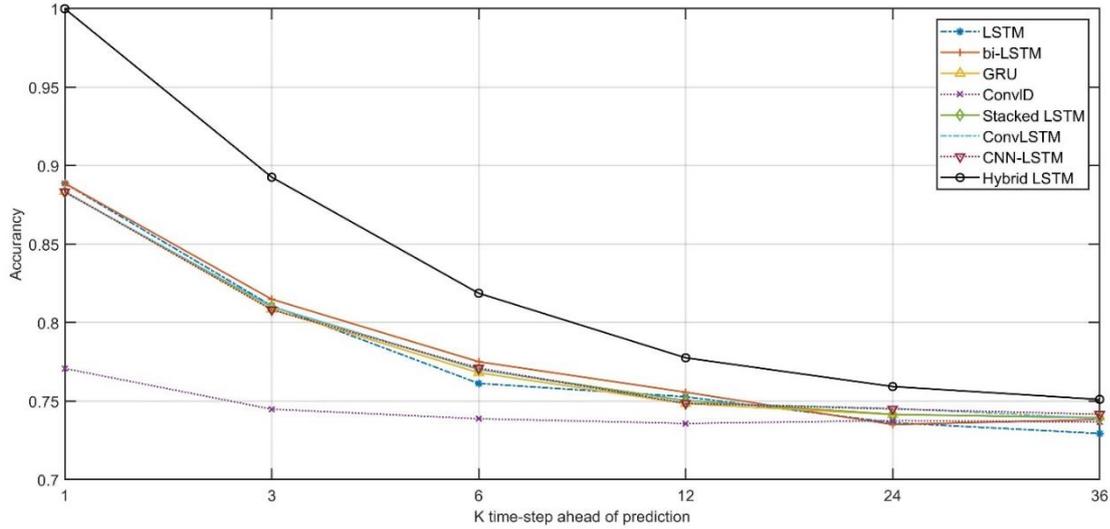

Figure 12. Comparison of the prediction accuracies of different DL models for multiple time-step ahead predictions.

## 4.3. Sensitivity analysis

To further explore the influence of the model parameters, we conduct a series of sensitivity analyses concerning five key model parameters, as shown in Table 8. Each experiment differs by varying the values of a tested hyperparameter while keeping other hyperparameters identical. The reported results are based on the average of 5 runs on the test dataset for all rapid chargers for 6-time-step prediction. Table 8 shows the summary of the accuracy performance for each tested hyperparameter value. The results show that using different hyperparameter settings could improve marginally the prediction accuracy which is consistent with our previous study (Schwemmle and Ma, 2021). We find that using three fully connected layers allows the enhanced learning of non-linear relationships from the large (i.e.147) feature vector. For the kernel size of the LSTM block, we find it performs better with a size between 18 and 54. The best dropout rate is 0.2, and the best-performing number of training epochs is 15 to avoid overfitting on the test dataset (Figure 13). For new charging occupancy datasets (e.g. newly available data from existing or new charging stations), one can tune the hyperparameters by the proposed sequential tuning approach using the values (Table 3) found in this study as a reference and test limited candidate values for each parameter around this reference point to adapt new scenarios quickly.

Table 8. Influence of the hyperparameters of the hybrid LSTM model on prediction accuracy.

| Hyperparameter | Value | Accuracy on the test dataset |
|---|---|---|
| Learning rate | [0.0005,**0.001**,0.002,0.004,0.008,0.012] | 0.8186 **0.8193** 0.8187 0.8178 0.8149 0.8110 |
| Number of fully connected layers* | [1,2,**3**] | 0.8156 0.8184 **0.8201** |
| Kernel size of the LSTM block | [18,**36**,54,72,108,144] | 0.8187 **0.8201** 0.8199 0.8183 0.8185 0.8188 |
| Dropout | [0.1, 0.2, 0.3, 0.4, 0.5] | 0.8188 **0.8198** 0.8185 0.8188 0.8166 |

| | | | | | | |
|---|---|---|---|---|---|---|
| Number of training epochs | [10,**15**,20,25,30,40, 50,60,70,80,90,100] | 0.8185 0.8187 0.8137 | **0.8195** 0.8179 0.8123 | 0.8193 0.8166 | 0.8191 0.8163 | 0.8195 0.8156 | |

Remark: *Top-right branch of the hybrid LSTM network architecture.

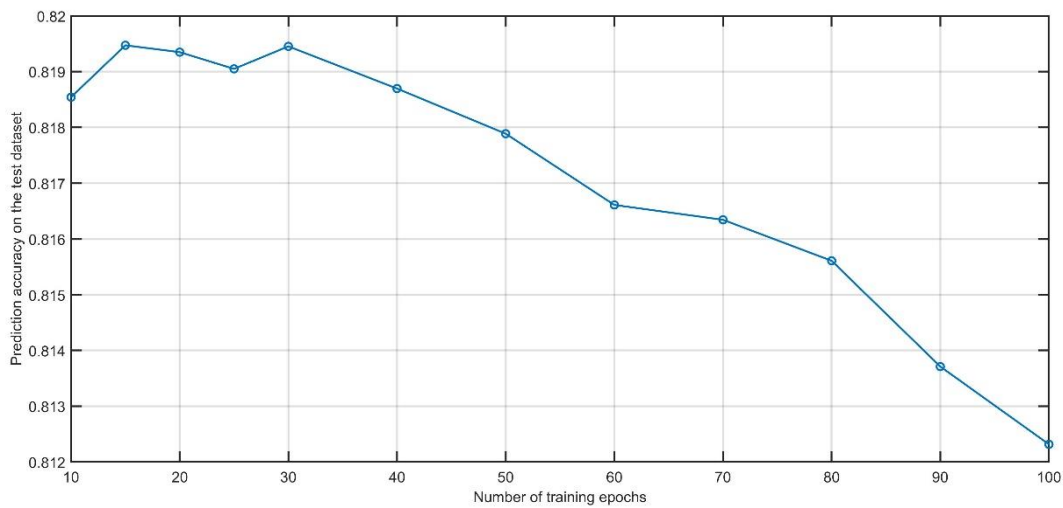

Figure 13. Influence of the number of training epochs on the prediction accuracy of the test dataset.

## 5. Discussion and conclusions

This paper proposes a new approach for predicting the occupancy of EV charging stations. This problem is really important for the management of EV fleets and has hardly been addressed in the scientific literature as a discrete EV charging occupancy-state modelling problem. To do so, we propose a hybrid LSTM neural network that considers both short-term and long-term charging occupancy states to model EV charging occupancy profiles at chargers. An open dataset provided by the city of Dundee, UK, is used as a basis to implement our approach and verify its performance. This method is compared with four other more conventional machine learning methods and three other DL networks. In all cases, the accuracy rate and F1 score show higher performance, both for short-term prediction (10 minutes: +22% improvement in F1 score over the best competing approach) and long-term prediction (6 hours: +2%). Similar findings are obtained when comparing other state-of-the-art deep learning approaches. The computational codes and data are freely available for their potential applications and extensions.

These results show a strong potential for the improvement of charging station occupancy prediction methods, which allows EV-based mobility service operators to develop smart-charging scheduling strategies. Moreover, the proposed methodology could lead to a more advanced recommendation or allocation strategies than what exists today—for example, using multi-objective optimisation approaches to meet various constraints (e.g. which charging station should a user consider given waiting time, potential new arrivals, and the geographic position of that station). Similarly, the practical development of these new strategies would require a high-speed exchange of information and a full, low-latency interconnected network—which could involve distributed network issues or ones specific to the wireless and 5G communication literature. Future work might extend the proposed methodology for other time series forecasting involving continuous variables with heterogeneous (time series and cross-sectional) data. Other possible directions might involve different mixed architectures with additional spatiotemporal features for different applied fields such as EV energy consumption demand and taxi demand arrival pattern forecasting.

**Acknowledgements**

The work was supported by the Luxembourg National Research Fund (C20/SC/14703944).

**Data Availability**

The data and Python codes used in this study are available at https://github.com/tym2021.

networks. *Energy*, *182*, 72–81.
19. Kiperwasser, E., Goldberg, Y. (2016). Simple and accurate dependency parsing using bidirectional LSTM feature representations. *Transactions of the Association for Computational Linguistics*, *4*, 313–327.
20. Laib, O., Khadir, M.T., Mihaylova, L. (2019). Toward efficient energy systems based on natural gas consumption prediction with LSTM Recurrent Neural Networks. *Energy*, *177*, 530–542.
21. Lee, Z.J., Li, T., Low, S.H. (2019). ACN-data: Analysis and applications of an open EV charging dataset. In *Proceedings of the Tenth ACM International Conference on Future Energy Systems*, 139–149.
22. Linoff, G.S., Berry, M.J.A. (2011). D*ata Mining Techniques for Marketing, Sales and Customer Support*. Wiley.
23. Ma, T.-Y., Pantelidis, T., Chow, J.Y. (2019). Optimal queueing-based rebalancing for one-way electric carsharing systems with stochastic demand. Paper presented in Transportation Research Board 98th Annual Meeting. https://arxiv.org/abs/2106.02815.
24. Ma, T.-Y. (2021). Two-stage battery recharge scheduling and vehicle-charger assignment policy for dynamic electric dial-a-ride services. *PLOS ONE*, 16(5), p.e0251582–e0251582.
25. Ma, T.-Y., Xie, S. (2021). Optimal fast charging station locations for electric ridesharing with vehicle-charging station assignment. Transportation Research Part D: Transport and Environment, 90, 102682.
26. Majidpour, M., Qiu, C., Chu, P., Pota, H. R., Gadh, R. (2016). Forecasting the EV charging load based on customer profile or station measurement?. Applied energy, 163, 134-141.
27. Motz, M., Huber, J. Weinhardt, C., (2021). Forecasting BEV charging station occupancy at work places. In: Reussner, R. H., Koziolek, A. and Heinrich, R. (Hrsg.), INFORMATIK 2020. Gesellschaft für Informatik, Bonn. (S. 771-781). DOI: 10.18420/inf2020_68
28. Pantelidis, T., Li, L., Ma, T.Y., Chow, J.Y., Jabari, S.E. (2020). Node-charge graph-based online carshare rebalancing with capacitated electric charging. *arXiv:2001.07282*.
29. Pascanu, R., Gulcehre, C., Cho, K., Bengio, Y. (2013). How to construct deep recurrent neural networks. arXiv preprint arXiv:1312.6026.
30. Petersen, N. C., Rodrigues, F., Pereira, F. C. (2019). Multi-output bus travel time prediction with convolutional LSTM neural network. Expert Systems with Applications, 120, 426-435.
31. Sajjad, M., Khan, Z. A., Ullah, A., Hussain, T., Ullah, W., Lee, M. Y., Baik, S.W. (2020). A novel CNN-GRU-based hybrid approach for short-term residential load forecasting. *IEEE Access*, *8*, 143759–143768.
32. Sawers, P. 2019. Google Maps will now show real-time availability of electric vehicle charging stations. https://venturebeat.com/2019/04/23/google-maps-will-now-show-real-time-availability-of-charging-stations-for-electric-cars/
33. Schwemmle, N. (2021). Short-term spatio-temporal demand pattern predictions of trip demand, Master Thesis, Katholieke Universiteit Leuven. https://zenodo.org/record/4514435#.YRZTNYgzbIU
34. Schwemmle, N., Ma, T.-Y. (2021). Hyperparameter Optimization for Neural Network based Taxi Demand Prediction. In: Proceedings of the BIVEC-GIBET Transport Research Days 2021.
35. Shi, X. , Chen, Z. , Wang, H. , Yeung, D.-Y. , Wong, W.-K. , Woo, W.-C. (2015). Convolutional LSTM network: A machine learning approach for precipitation nowcasting. In Proceedings of the 28th international conference on neural information processing systems –Volume 1, pp. 802–810.
36. Smola, A.J., Schölkopf, B. (2004). A tutorial on support vector regression. *Statistics and Computing. 14*(3), 199–222.
37. Soldan, F., Bionda, E., Mauri, G., Celaschi, S. (2021). Short-term forecast of EV charging stations occupancy probability using big data streaming analysis. arXiv:2104.12503.
38. Tian, Z., Jung, T., Wang, Y., Zhang, F., Tu, L., Xu, C., Tian, C., Li, X.Y. (2016). Real-Time Charging Station Recommendation System for Electric-Vehicle Taxis. IEEE Trans. Intell. Transp. Syst., 17, 3098– 3109. https://doi.org/10.1109/TITS.2016.2539201
39. Ullah, F.U.M., Ullah, A., Haq, I.U., Rho, S., Baik, S.W. (2019). Short-term prediction of residential power energy consumption via CNN and multi-layer bi-directional LSTM networks. *IEEE Access*, *8*, 123369–123380.
40. Van Houdt, G., Mosquera, C., Nápoles, G. (2020). A review on the long short-term memory model. *Artif. Intell. Rev., 53*, 5929–5955.
41. Verma, A., Asadi, A., Yang, K., Maitra, A., Asgeirsson, H. (2019). Analyzing household charging patterns of Plug-in electric vehicles (PEVs): A data mining approach. *Computers & Industrial*

*Engineering*, *128*, 964–973.
42. Wang, J.Q., Du, Y., Wang, J. (2020). LSTM based long-term energy consumption prediction with periodicity. *Energy*, *197*, 117197.
43. Yang, Y., Hong, W., Li, S. (2019). Deep ensemble learning based probabilistic load forecasting in smart grids. *Energy*, *189*, 116324.
44. Yuan, Y., Zhang, D., Miao, F., Chen, J., He, T., Lin, S., (2019). P2Charging: Proactive partial charging for electric taxi systems. In: *Proceedings - International Conference on Distributed Computing Systems*. https://doi.org/10.1109/ICDCS.2019.00074
45. Zhang, Q., Wang, H., Dong, J., Zhong, G., Sun, X. (2017). Prediction of sea surface temperature using long short-term memory. *IEEE Geoscience and Remote Sensing Letters*, 14(10), 1745-1749.
46. Zhao, K., Khryashchev, D., Vo, H. (2019). Predicting taxi and uber demand in cities: Approaching the limit of predictability. IEEE Transactions on Knowledge and Data Engineering, 33(6), 2723-2736.
20